\documentclass[runningheads]{llncs}
\usepackage{graphicx}
\usepackage{caption}
\usepackage{subcaption}
\begin{document}
\title{Naturalistic Robot Arm Trajectory Generation via Representation Learning}
%
%
\author{Jayjun Lee \and
Adam J. Spiers\orcidID{0000-0002-3221-1000}}
\authorrunning{J. Lee and A. Spiers.}
%
\institute{Imperial College London, London, UK\\
\email{\{jayjun.lee19,a.spiers\}@imperial.ac.uk}}
\maketitle              

\begin{abstract}
The integration of manipulator robots in household environments suggests a need for more predictable and human-like robot motion. This holds especially true for wheelchair-mounted assistive robots that can support the independence of people with paralysis. One method of generating naturalistic motion trajectories is via the imitation of human demonstrators. This paper explores a self-supervised imitation learning method using an autoregressive spatio-temporal graph neural network for an assistive drinking task. We address learning from diverse human motion trajectory data that were captured via wearable IMU sensors on a human arm as the action-free task demonstrations. Observed arm motion data from several participants is used to generate natural and functional drinking motion trajectories for a UR5e robot arm. 

\end{abstract}


\section{Introduction}
For people with motion impairments, the ability to feed oneself is a major factor of independence \cite{Chung2013FunctionalReview}. Recently, wheelchair- or desk-mounted robotic manipulators have been implemented with these tasks in mind \cite{Beaudoin2018ImpactsReview,Chi2019RecentArm}.  In human-robot interaction (HRI), it has been observed that the human comfort and confidence may be increased by generating predictable and naturalistic motion paths \cite{Gulletta2020Human-LikeReview,Spiers2016BiologicallyApproaches}. As such, we are aiming to add human-like arm motion to an assistive drinking task.




To generate human-like robot arm motion we collect human arm movement data using wearable IMUs. We then reconstruct action-free human arm trajectories to gain access to low-dimensional states, and use an autoregressive spatio-temporal graph neural network (GNN) to ingest this data in a self-supervised way. We learn internal model representations of human drinking dynamics that exploit the spatial and temporal relation between arm joints based on the Space-Time Separable Graph Convolutional Network (STS-GCN) \cite{Sofianos2021Space-Time-SeparableForecasting}. By behaviour cloning (BC) from the human motion data collected via IMUs, we were able to generate diverse, human-like drinking robot arm motion that is functional across various bottle positions with heuristics to complete other subtasks in sequence.

In this work we have adapted the STS-GCN architecture from the human pose prediction community into an autoregressive GNN for self-supervised imitation learning for robotics, with the Mean Per Joint Position Error (MPJPE) as the BC loss. As a result, the new system learns an internal model dynamics of naturalistic drinking motion with relatively sparse input data, making it suited to fewer (motion captured) demonstrations. The more compact result is also better suited for implementation on physical hardware with predictions further back in time making it suited to functional tasks.

\vspace{-0.5cm}
\begin{figure}
\centering
\includegraphics[width=1.0\linewidth]{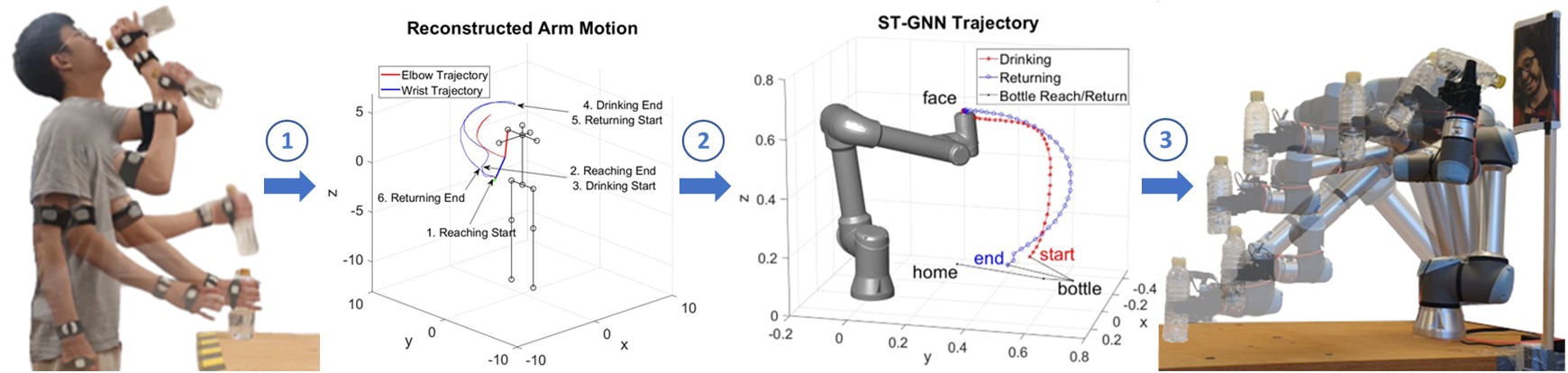}
\caption{Human drinking motion is captured using wearable IMUs, with the arm trajectory reconstructed to form an action-free demonstration. An autoregressive spatio-temporal GNN learns the motion dynamics from diverse drinking data to generate generalised naturalistic drinking motions which are scaled for a UR5e.}
\label{summary figure}
\end{figure}

\vspace{-1.2cm}
\section{Related Work}

\subsection{Human-like Arm Motion Generation for Robots}
It has been proposed that human-like behaviours of robotic manipulators can ensure safety, predictability, and social acceptance \cite{Gulletta2020Human-LikeReview,Spiers2016BiologicallyApproaches}. Many research efforts have aimed for various aspects of human-like robot motion planning. One popular approach is movement primitives that decompose motion into a set of primitives that can be combined to generate complex movements and learned from human demonstrations \cite{paraschos2018using}. We have adopted a self-supervised learning method that can generate diverse and generalisable human-like motion while learning an internal model dynamics with an autoregressive structure and without primitives. It is noted that our approach and MPs could potentially complement each other.

Human motion forecasting deals with the problem of predicting the 3D coordinates of $V$ body joints for the future $K$ frames, given past $T$ frames. A skeleton-based model of human body may be used to form a graph structure, where each joint is a node \cite{Jain2016Structural-RNN:Graphs,Mao2020HistoryAttention}. In \cite{Sofianos2021Space-Time-SeparableForecasting}, the STS-GCN model is introduced, which learns to encode the human body dynamics by factorising the spatio-temporal graph adjacency matrix to separate spatial and temporal adjacency matrices and focus on the joint-joint and temporal relations. We modify this architecture to learn from relatively sparsely logged data by extending the model to train autoregressively with a self-supervised loss. The result is a more compact learned internal model of human motion dynamics that can predict much further in time. We also take an embodied approach that maps the generated human arm trajectory onto a real robot arm to complete functional tasks, as opposed to visualisations of simulated skeleton models. Unlike in \cite{Sofianos2021Space-Time-SeparableForecasting}, the input trajectory segment to our system is not the initial frames of a continuous action, but rather a preparatory motion to reach and grasp a bottle prior to the generated movement of bringing the bottle to a user's mouth. 

\vspace{-0.8cm}
\section{Methods}
\vspace{-0.4cm}
To collect human drinking motion data, 3 MetaMotionS+ IMUs \textit{(MBientLab)} are attached using Velcro straps along the participant's right arm on the upper arm, forearm, and back of hand. Euler angles are logged at 100Hz and preprocessed to address discontinuities and noise. Five participants (two female, mean age of 23.2 years) each provided 10 drinking demonstrations for 6 discrete bottle positions on a 2-by-3 grid on a desk. Each recorded trajectory is discretely down-sampled to 150 samples and split into reaching, drinking and returning phases in Fig. \ref{summary figure}.

\begin{figure}
    \centering
    \includegraphics[trim={0 2.5cm 0 0},width=1\textwidth]{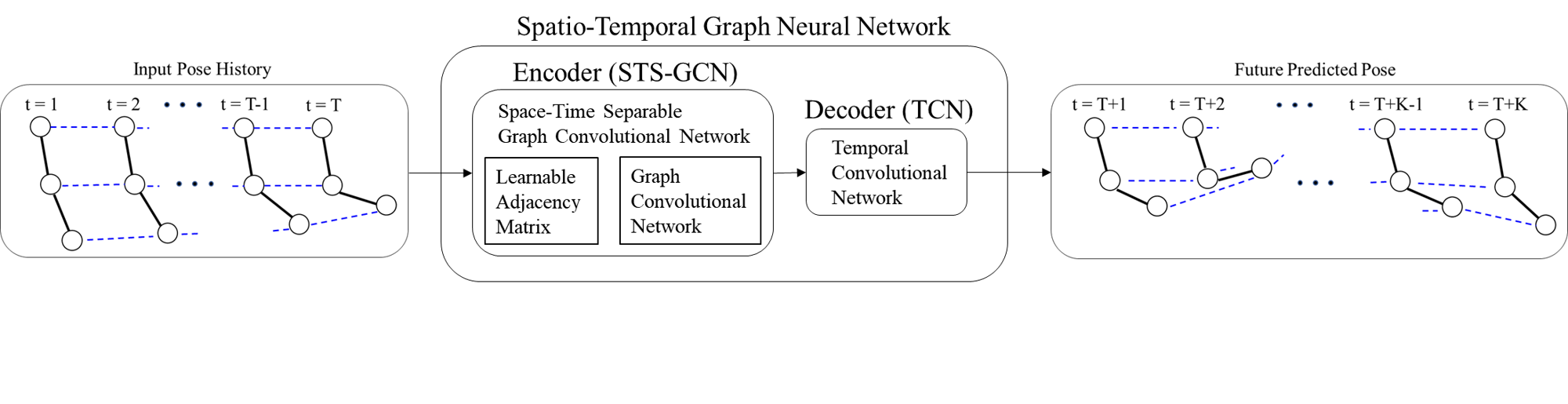}
    \vspace{-0.5cm}
    \caption{An overview of the autoregressive GNN adapted from the STS-GCN.}
    \label{my arch of stsgnn}
\end{figure}

The GNN is shown in Fig. \ref{my arch of stsgnn}. The encoder has four STS-GCN layers with the input graph of $\small{T=30}$, which learns the adjacency matrix of the input to highlight certain space-time edges with feature graphs. The decoder has five TCN layers to generate the output graph of $\small {K=30}$ with 3D joint coordinates. The learned graph representations act as the internal model for the drinking dynamics to generate the subsequent motion segment given the input segment. The model is trained to minimise the MPJPE loss in Eq. \ref{mean per joint position error} between the autoregressively generated 120-frame drinking trajectory and the ground-truth self-supervised label. This requires a recursive forwarding of its output to its input four times.

\vspace{-0.2cm}

\begin{equation}
    L_{MPJPE} = \frac{1}{VK}
    \sum_{k=T+1}^{T+K}\sum_{v=1}^{V}\left \| \hat{x}_{vk}-x_{vk} \right \|_2
    \label{mean per joint position error}
\end{equation}

$x_{vk}, \hat{x}_{vk} \in {R}^3$ are the true and predicted joint $v$ positions at frame $k$. $V$ is the number of nodes per frame. $T$ and $K$ are the number of input and output frames.

As the generated human trajectory resides in the human workspace, we map and linearly scale the human wrist 3D trajectory to the robot arm's end-effector (EE) workspace, safely reaching the user's mouth. Future work would integrate sensing solutions to deal with the user moving and force-interactions.

\vspace{-0.4cm}
\vspace{-1cm}


\section{Results}

\begin{figure}
\makebox[\linewidth][c]{%
    \begin{subfigure}[b]{.4\textwidth}
    \centering
    \includegraphics[height=4cm]{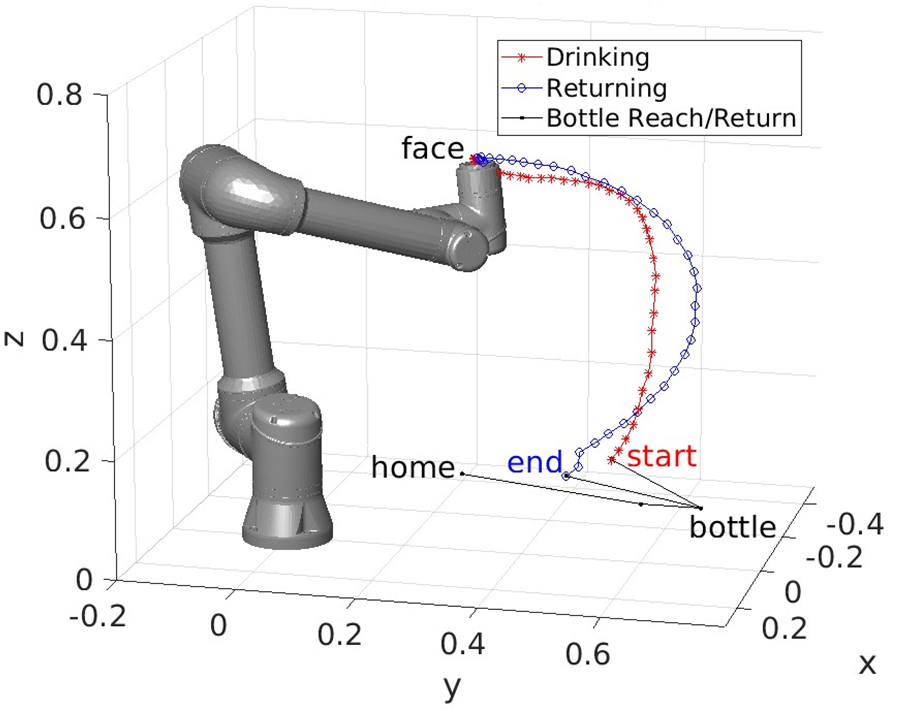}
    \caption{GNN trajectory.}
    \end{subfigure}%
    \begin{subfigure}[b]{.4\textwidth}
    \centering
    \includegraphics[height=4cm]{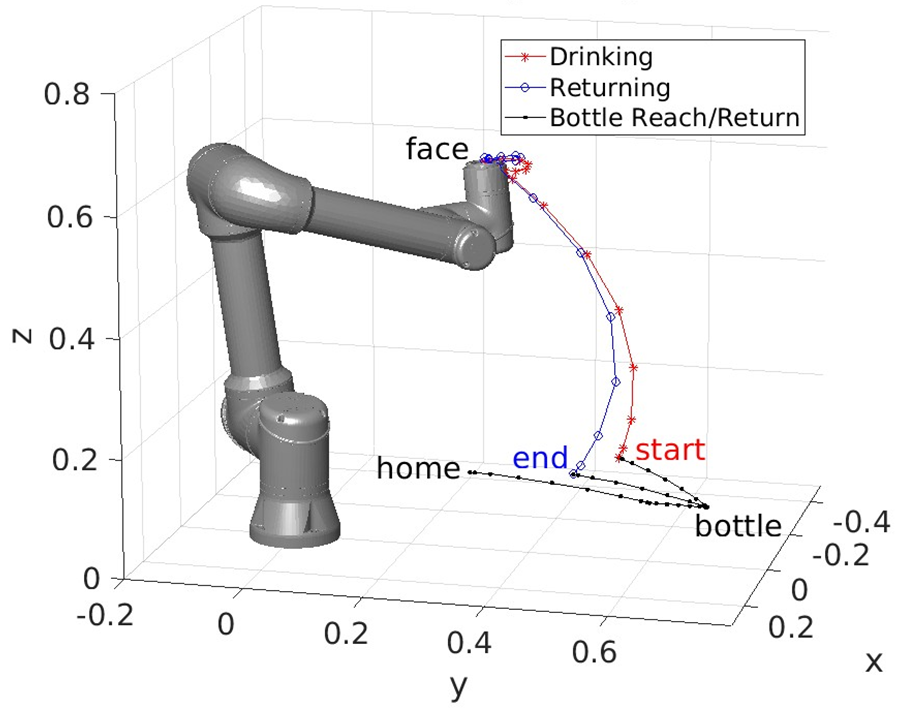}
    \caption{Joint-space trajectory.}
    \end{subfigure}
}
\caption{A comparison between a drinking trajectory generated by the GNN (a) and by joint-space inverse kinematic interpolation with trapezoidal velocity profile. (b) Note that a classic task space IK trajectory would be a straight line.}
\label{compare stgnn and jointspace}
\vspace{-0.4cm}
\end{figure}

\vspace{-0.2cm}
A 6 DOF UR5e robot arm was used with a parallel jaw gripper adapted from the ROBOTIS Open-Manipulator X robot, Fig. \ref{summary figure}. In Fig. \ref{compare stgnn and jointspace} we compare the GNN trajectory with a typical joint-space IK trajectory. Pronounced curves with hysteresis are present in the GNN trajectory. Such hysteresis also appears in human reaching motions \cite{Spiers2016BiologicallyApproaches}. We also test our trained model on unseen test bottle positions placed within the aforementioned 2-by-3 grid of bottle positions.

\section{Conclusion and Future work}
\vspace{-0.2cm}

We have proposed a preliminary GNN-based self-supervised imitation learning framework, using human demos to generate human-like robot arm drinking motion from a reach-to-grab motion. In future work, this could be extended by multi-task learning and with a camera to observe scene obstacles for more Activities of Daily Living where human-like motion is beneficial for assistive robots.

\vspace{-0.4cm}

\bibliographystyle{splncs04}
\bibliography{ref_manual}

\begin{thebibliography}{1}
\providecommand{\url}[1]{\texttt{#1}}
\providecommand{\urlprefix}{URL }
\providecommand{\doi}[1]{https://doi.org/#1}

\bibitem{Beaudoin2018ImpactsReview}
Beaudoin, M., Lettre, J., Routhier, F., Archambault, P.S., Lemay, M.,
  G{\'{e}}linas, I.: {Impacts of Robotic Arm Use on Individuals with Upper
  Extremity Disabilities: A Scoping Review}. Canadian Journal of Occupational
  Therapy  (2018)

\bibitem{Chi2019RecentArm}
Chi, M., Yao, Y., Liu, Y., Zhong, M.: {Recent Advances on Human-Robot Interface
  of Wheelchair-Mounted Robotic Arm}. Recent Patents on Mechanical Engineering
  (2019)

\bibitem{Chung2013FunctionalReview}
Chung, C.S., Wang, H., Cooper, R.A.: {Functional assessment and performance
  evaluation for assistive robotic manipulators: Literature review}. Journal of
  Spinal Cord Medicine  (2013)

\bibitem{Gulletta2020Human-LikeReview}
Gulletta, G., Erlhagen, W., Bicho, E.: {Human-Like Arm Motion Generation: A
  Review}. Robotics  (2020)

\bibitem{Jain2016Structural-RNN:Graphs}
Jain, A., Zamir, A.R., Savarese, S., Saxena, A.: {Structural-RNN: Deep learning
  on spatio-temporal graphs}. IEEE Conference on Computer Vision and Pattern
  Recognition (CVPR)  (2016)

\bibitem{Mao2020HistoryAttention}
Mao, W., Liu, M., Salzmann, M.: {History Repeats Itself: Human Motion
  Prediction via Motion Attention}. In: European Conference on Computer Vision
  (ECCV) (2020)

\bibitem{paraschos2018using}
Paraschos, A., Daniel, C., Peters, J., Neumann, G.: Using probabilistic
  movement primitives in robotics. Autonomous Robots  \textbf{42},  529--551
  (2018)

\bibitem{Sofianos2021Space-Time-SeparableForecasting}
Sofianos, T., Sampieri, A., Franco, L., Galasso, F.: {Space-Time-Separable
  Graph Convolutional Network for Pose Forecasting}. In: IEEE International
  Conference on Computer Vision (ICCV) (2021)

\bibitem{Spiers2016BiologicallyApproaches}
Spiers, A., Khan, S.G., Herrmann, G.: {Biologically inspired control of
  humanoid robot arms: Robust and adaptive approaches}. Springer International
  Publishing (2016)

\end{thebibliography}
\end{document}